\documentclass[10pt,twocolumn,letterpaper]{article}

\usepackage[pagenumbers]{iccv} 
\PassOptionsToPackage{table,dvipsnames}{xcolor}  
\usepackage{iccv}
\usepackage{times}
\usepackage{epsfig}
\usepackage{graphicx}
\usepackage{amsmath}
\usepackage{amssymb}
\usepackage{bbding}
\usepackage{pifont}
\usepackage{wasysym}

\usepackage{booktabs} 
\usepackage[most]{tcolorbox} 

\usepackage{fontawesome}
\definecolor{iccvblue}{rgb}{0.21,0.49,0.74}
\usepackage[pagebackref,breaklinks,colorlinks,allcolors=iccvblue]{hyperref}

\newcommand{\shortname}{X-Fusion }

\usepackage{ragged2e} 
\definecolor{backgroundcolor}{HTML}{F5F5F5}
\definecolor{borderline}{HTML}{D0D0D0}
\definecolor{tablebg}{HTML}{F5F5F5}

\newtcolorbox{findingbox}{
    sharpish corners, 
    boxrule = 0pt,
    leftrule = 3pt, 
    enhanced,
    fuzzy shadow = {0pt}{-2pt}{-0.5pt}{0.5pt}{black!35}, 
    colback=backgroundcolor, 
    colframe=borderline,  
    boxsep=2pt, 
    left=5pt, 
    right=5pt, 
    before upper={\noindent\justifying} 
}
\def\paperID{1796} 
\def\confName{ICCV}
\def\confYear{2025}

\usepackage{xcolor}   
\usepackage{tikz}     

\usepackage{gradient-text}
\title{X-Fusion: Introducing New Modality to Frozen Large Language Models}
\author{
Sicheng Mo$^{1}$ \hspace{0.5mm}
Thao Nguyen$^{2}$ \hspace{0.5mm}
Xun Huang$^{3}$ \hspace{0.5mm}
Siddharth Srinivasan Iyer$^{3}$ \hspace{0.5mm}
Yijun Li$^{3}$ \hspace{0.5mm}
Yuchen Liu$^{3}$ \\
Abhishek Tandon$^{3}$ \hspace{0.5mm}
Eli Shechtman$^{3}$ \hspace{0.5mm}
Krishna Kumar Singh$^{3}$ \hspace{0.5mm}
Yong Jae Lee$^{2}$ \hspace{0.5mm}
Bolei Zhou$^{1}$ \hspace{0.5mm}
Yuheng Li$^{3}$\\[1mm]
$^{1}$University of California, Los Angeles \hspace{0.5mm}
$^{2}$University of Wisconsin–Madison \hspace{0.5mm}
$^{3}$Adobe Research
\\[1mm]
{\normalsize \url{https://sichengmo.github.io/XFusion/}}
}

\begin{document}
\maketitle
\begin{abstract}

We propose X-Fusion, a framework that extends pretrained Large Language Models (LLMs) for multimodal tasks while preserving their language capabilities.
X-Fusion employs a dual-tower design with modality-specific weights, keeping the LLM’s parameters frozen while integrating vision-specific information for both understanding and generation.
Our experiments demonstrate that X-Fusion consistently outperforms alternative architectures on both image-to-text and text-to-image tasks.
We find that incorporating understanding-focused data improves generation quality, reducing image data noise enhances overall performance, and feature alignment accelerates convergence for smaller models but has minimal impact on larger ones.
Our findings provide valuable insights into building efficient unified multimodal models.

\end{abstract}


\vspace{-2mm}
\section{Introduction}

\begin{figure}[t]
    \centering
    \includegraphics[width=1\linewidth]{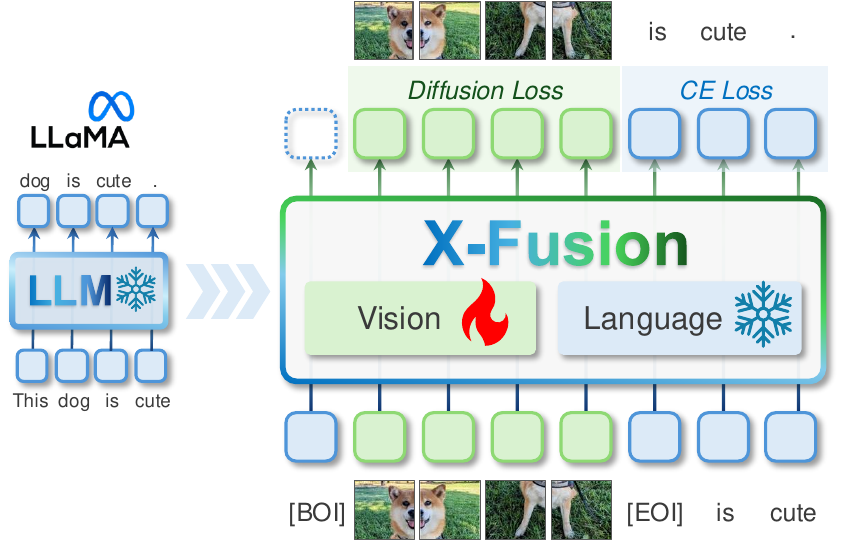}
    \vspace{-5mm}
    \caption{
    We introduce \textbf{X-Fusion} - a novel framework that adapts pretrained LLMs (e.g., LLaMA) to new modalities (e.g., vision) while retaining their language capabilities and world knowledge.
}

    \label{fig:x-fusion}
    \vspace{-2mm}
    
\end{figure}

Large Language Models (LLMs)~\cite{gpt,gpt2,gpt3,t5,gopher,flan,chinchilla,palm,palm2,jiang2024mixtralexperts,jiang2023mistral7b,deepseekllm,gemma} have not only achieved unprecedented capabilities for language processing tasks (e.g., conversational AI~\cite{vicuna,llama2,phi3,llama3,qwen,cai2024internlm2,deepseekv2}),
but also emerged as foundation tools to solve multiple language-related challenges (e.g., coding~\cite{codex,codegen,starcoder}).
However, as humans, we do not communicate solely through text, but also extend to other modalities, such as vision.
For example, instead of simply saying ``This dog is cute'', we might show a photo of the dog to enhance the message: ``[image] is cute'' (Fig.~\ref{fig:x-fusion}).
Thus, a truly versatile AI model should not only understand, reason, and generate textual output, but also must have abilities to understand, reason, and generate visual information.
Moreover, these models should be unified to process and generate language and vision simultaneously, creating a more comprehensive interactive experience.

\begin{figure*}[th]
\centering 
\includegraphics[width=1\textwidth,page=2]{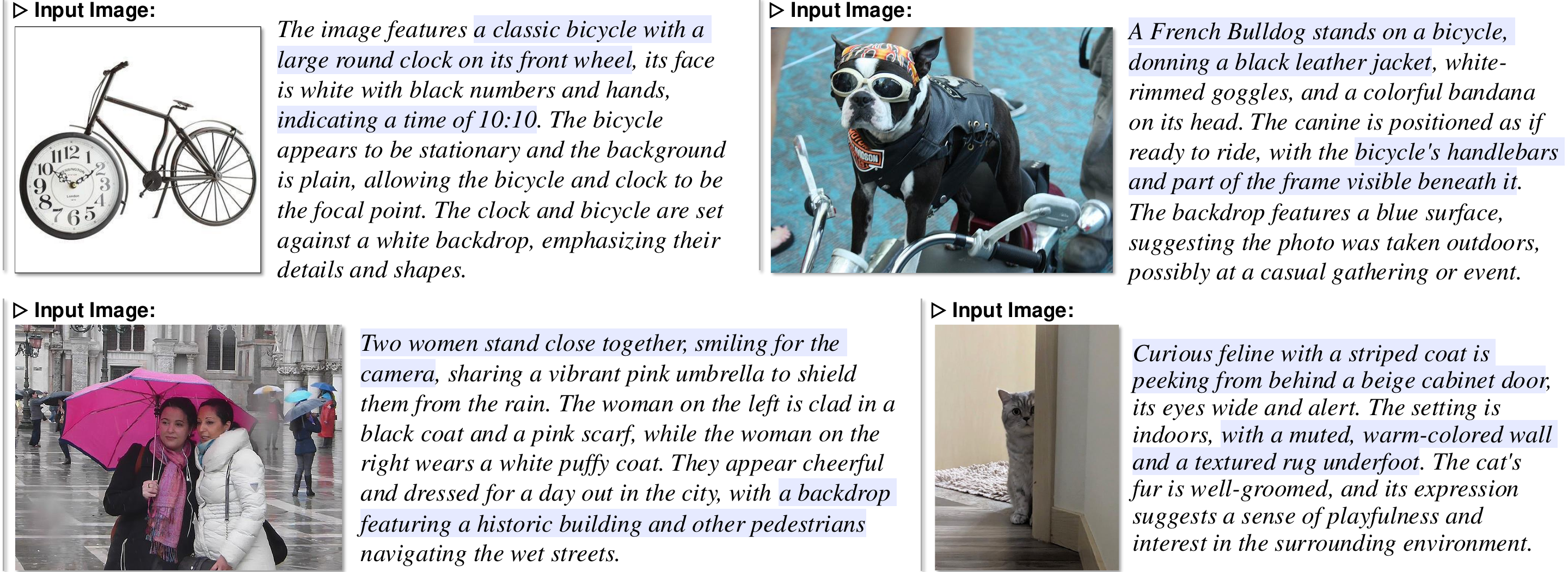}
    \vspace{-4mm}
    \caption{
    \textbf{Captions generated by X-Fusion} demonstrate high details and strong visual alignment with the image inputs. 
    }
    \vspace{-3mm}
\label{fig: qualitative results}
\end{figure*}
%
%
To achieve a unified model, some approaches focus on training unified vision-language models entirely from scratch using next-token prediction loss~\cite{CM3Leon,gemini,Chameleon,emu3}. A recent alternative, Transfusion~\cite{transfusion}, adopts a domain-specific strategy by combining next-token prediction loss for language with diffusion loss for images. This hybrid architecture has significantly advanced performance, demonstrating greater promise than purely autoregressive approaches like Chameleon~\cite{Chameleon}. However, training such models from scratch demands immense computational resources (e.g.,~\cite{transfusion} trained on 2T tokens) and necessitates full retraining for each new modality.
Given these shortcomings, another prominent research direction explores how to reuse powerful pretrained LLMs and introduce vision abilities to them~\cite{dreamllm,gill,tong2024metamorph}, offering a more practical and efficient way for unified multimodal model training.

Research on adapting LLMs~\cite{vicuna,llama2,phi3,llama3,qwen} with image understanding has shown promising results through ``visual instruction tuning''~\cite{liu2023llava,liu2023improvedllava,zhu2024minigpt,chen2024internvl}. These models typically fine-tune the LLM to align the text feature space with pretrained vision encoders (e.g., CLIP~\cite{clip}), thus, might degrade original language capabilities~\cite{zhai2023investigating,forgetting2}. 
Unlike image understanding, image generation poses greater difficulties, as it demands output capabilities in a new feature space.
A large body of work~\cite{dreamllm,gill,Emu2,seed-x,seed-llama,tang2023codi2,januspro} tackled this issue by leveraging pretrained image generation models (e.g., Stable Diffusion~\cite{stablediffusion}).
However, as these frameworks are not unified, this approach creates several limitations: limited cross-modal reasoning, restricted in-context learning, and increased error accumulation~\cite{2017multimodalmachinelearningsurvey}.
Most critically, these approaches typically require fine-tuning the LLM backbone, degrading inherited text generation ability~\cite{zhai2023investigating,forgetting2}.
This raises fundamental research question: \emph{Is there a better way to introduce new modalities to pretrained LLMs?}

Motivated by these observations, we propose \textbf{X-Fusion}, a new approach that addresses two challenges: (i) retaining the language abilities of the pre-trained LLM while (ii) adapting it with image generation capabilities.
First, X-Fusion freezes all language weights, denoted as the {\em text tower}, thus preserving the inherent language abilities. Second, instead of fine-tuning the LLM, we introduce a {\em vision tower} with separate vision weights in each layer to help process visual information for the LLM (Fig~\ref{fig:x-fusion}).
This approach aligns text and vision features not only at input or output level, but also at the intermediate processing level.
It is worth noting that this architecture is flexible in terms of design — the vision and text towers can have asymmetric architectures. Moreover, the framework naturally extends to additional modalities (e.g., audio) by introducing dedicated modality-specific towers, ensuring efficient and scalable multimodal integration while keeping each modality independent.

\begin{figure*}[h]
\centering 
\includegraphics[width=1\textwidth]{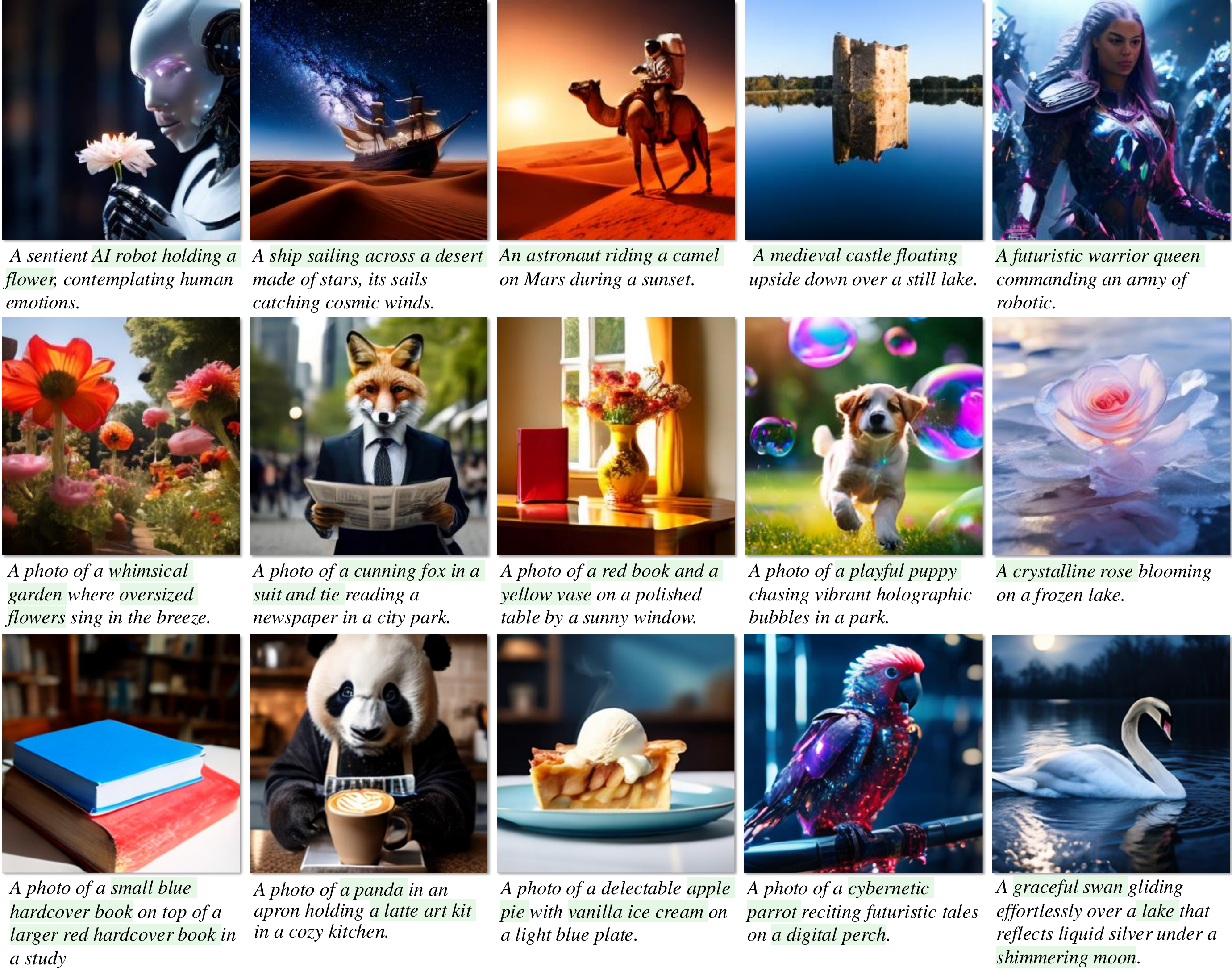}
   \vspace{-5mm}
    \caption{
    \textbf{Images generated by X-Fusion} demonstrate high visual quality and strong text alignment with the input prompts.
    }
   \vspace{-3mm}
\label{fig: qualitative results}
\end{figure*}

While architectural innovations are crucial for integrating visual capabilities into LLMs, nowadays, understanding the impact of training data is equally important. Thus, we conduct a comprehensive set of ablation studies from a data-centric perspective to build a scalable training strategy.
First, we note that creating image understanding samples without introducing noise to the images is crucial for training diffusion-based unified models. This approach enhances both image generation and understanding abilities due to more semantic visual representation learned from clean images. Based on this key component, we further observed cross task synergy---including more image understanding data that enhances generation performance. 
We also investigate the effectiveness of aligning our vision features with additional pre-trained representations. Our findings suggest that this extra alignment loss term may help smaller models converge faster, but the benefit diminishes as the model size increases.

In short, our contributions are: 
(i) X-Fusion - a novel framework that adapts pretrained LLMs to new modalities while retaining their language capabilities.
(ii) A systematic study on training strategy, offering insights for optimizing multi-modal learning.
(iii) Experimental results on both image-to-text and text-to-image tasks, validating the effectiveness of the proposed architecture.

\section{Related Work}
\noindent\textbf{Large Multimodal Models.}
The development of artificial intelligence has historically followed separate, modality-specific paths. Large Language Models (LLMs)~\cite{gpt,gpt2,gpt3,t5,gopher,flan,chinchilla,palm,palm2,jiang2024mixtralexperts,jiang2023mistral7b,deepseekllm,gemma} exclusively processed text input and generated textual responses, while computer vision models specialized in visual content understanding (e.g., object detection~\cite{yolo}) or visual generation (e.g., StyleGAN~\cite{stylegan}).
The emergence of vision encoder models like CLIP~\cite{clip} bridged text and image modalities, enabling two key advances: (1) Vision-language models that allow LLMs to ``see'' (e.g., LLaVA~\cite{liu2023llava}); and (2) Conditional visual generation models that can process textual input (e.g., Stable Diffusion~\cite{stablediffusion}).
Current research frontiers are focusing on integrating vision and language capabilities into unified models that can both process and generate multimodal content.
There are three main approaches: (1) Merging LLMs with pretrained image generation models (e.g., DreamLLM~\cite{dreamllm}, GILL~\cite{gill}); (2) Training LMMs via next-token prediction (e.g., Chameleon~\cite{Chameleon}); or (3) Training LMMs using both diffusion and next-token prediction losses (e.g., Transfusion~\cite{transfusion}).
Following the third approach, which has achieved state-of-the-art results across modalities, we instead propose initializing from frozen LLMs rather than training from scratch, significantly reducing computational costs and retaining the LLMs knowledge. 

\noindent\textbf{Leveraging Pretrained LLMs.} Since the success of LLMs~\cite{gpt,gpt2,gpt3,t5,gopher,flan,chinchilla,palm,palm2,jiang2024mixtralexperts,jiang2023mistral7b,deepseekllm,gemma}, researchers have discovered that pretrained LLMs can serve as effective baselines for various purposes (e.g., coding~\cite{codex,codegen,starcoder}).
Beyond being adapted for domain-specific tasks, prior works have shown that these pretrained models can also be fine-tuned to acquire new abilities, paving the way for a cost-effective approach to include more modalities like image understanding (e.g., LLaVA~\cite{liu2023llava}, Mini-GPT-4~\cite{zhu2024minigpt}), image generation (e.g., GILL~\cite{gill}, DreamLLM~\cite{dreamllm}, SEED-X~\cite{seed-x}), or both image understanding and generation (e.g., Show-o~\cite{show-o}, Emu3~\cite{emu3}, MetaMorph~\cite{tong2024metamorph}).
However, this method is not without limitations---often, when fine-tuning the LLMs' backbone, it risks compromising the original knowledge. To alleviate this shortcoming, in this work, we propose a novel method to extend vision-related abilities for LLMs while keeping all the language layers frozen, thus maintaining its original language abilities untouched. A concurrent work is LMFusion~\cite{shi2024llamafusion}, with a similar high-level approach. However, they use joint attention across text and vision tokens, which is less flexible than our proposed model design. 

\smallskip
\noindent\textbf{Task-specific Weights.} People have been exploring the use of specialized models tailored to different tasks in both the language~\cite{Shazeer2017OutrageouslyLN, Lepikhin2020GShardSG, Fedus2021SwitchTS, Du2021GLaMES} and vision domains~\cite{Riquelme2021ScalingVW, Li2021CollagingCG, Zhu2023ExploringSM}. The use of unshared parameters has also been explored in multi-modal settings.
For instance, CLIP~\cite{clip} and ImageBind~\cite{imagebind} focus on representation learning, while other works~\cite{Wang2022ImageAA, Wang2021VLMoUV, Shen2023ScalingVM} emphasize vision-conditioned language model pretraining.
However, these multi-modal approaches predominantly focus on visual understanding, neglecting generation tasks. Importantly, they are typically trained from scratch, which is computationally expensive.
A concurrent work, Playground-v3~\cite{Playground-v3}, takes a different approach by building upon LLMs, but freezes them to perform generation tasks only, without addressing visual understanding.

\begin{figure*}[t]
\centering 
\includegraphics[width=1\textwidth]{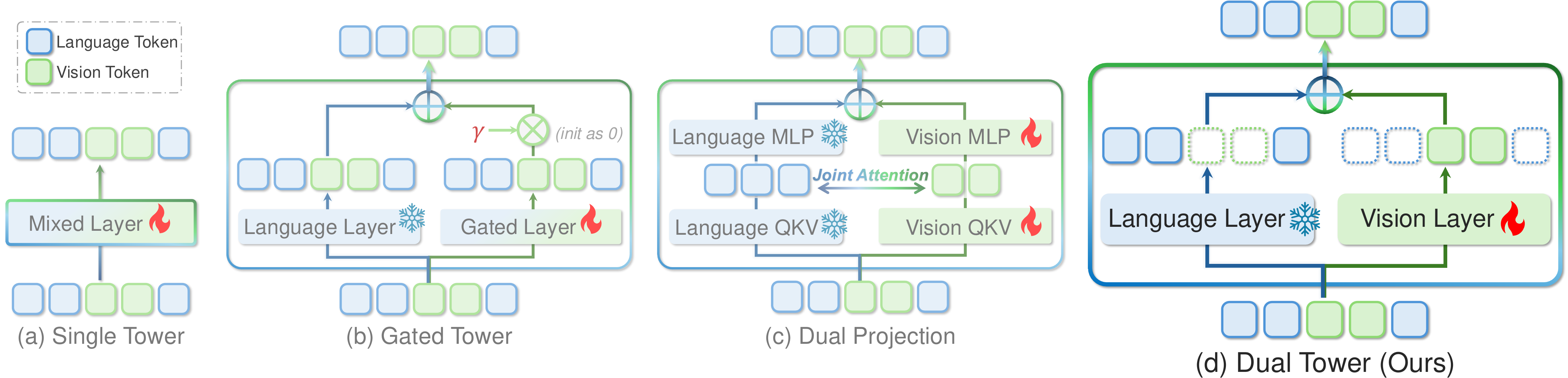}
    \caption{
    \textbf{Conceptual comparison of four model architecture baselines.} Here, we illustrate how each layer processes the sequential multi-modal feature. \textbf{(a) Single Tower:} Directly fine-tuning pre-trained LLM.
    \textbf{(b) Gated Layer: }Duplicate Each LLM layer as the gated vision layer, 
    \textbf{(c) Dual Projection:} Duplicate QKV matric and MLP layer for vision modality, 
    \textbf{(d) Dual Tower:} Duplicated transformer block for vision modality.
    }
\label{fig:pipeline comparaison}
\end{figure*}

\section{Preliminaries}

In this section, we provide a brief preliminary overview of the state-of-the-art recipe (proposed by~\cite{transfusion}) for training unified models in a hybrid manner, incorporating next-token prediction for language and diffusion models for image.

\subsection{Language Modeling via Autoregression}

LLMs are typically trained using an autoregressive modeling objective, where the joint probability of a sequence of language tokens $\mathbf{x}^{\text{txt}} = \{x^{\text{txt}}_1, x^{\text{txt}}_2, \dots, x^{\text{txt}}_N\}$ is factorized as a product of conditional probabilities:
\[
P_\theta(\mathbf{x}^{\text{txt}} \mid c) = \prod_{i=1}^{N} P_\theta(x^{\text{txt}}_i \mid x^{\text{txt}}_{<i}, c).
\]

Here, \( c \) represents optional conditioning information, which could include extracted features from other modalities (e.g., image representations) or task-specific context. However, in the standard setting, LLMs are usually trained without any additional conditioning (\( c \) is absent), and the predictions depend solely on the preceding tokens \( x^{\text{txt}}_{<i} \).

The training objective for this autoregressive model is to minimize the negative log-likelihood of the observed data:
\[
\mathcal{L}_{\text{AR}} = \mathbb{E}_{\mathbf{x}^{\text{txt}}, c} \left[ - \sum_{i=1}^{N} \log P_\theta(x^{\text{txt}}_i \mid x^{\text{txt}}_{<i}, c) \right].
\]

If conditioning information \( c \) is introduced (e.g., in multimodal or task-specific setups), it allows the model to extend its capabilities to tasks such as conditional text generation and cross-modal understanding.

\subsection{Image Modeling with Diffusion}
Diffusion models have emerged as one of the most successful approaches for image generation thanks to their ability to generate high-quality images.
Diffusion models are trained to gradually reverse the process of adding noise to an image, starting from a noise vector \( \mathbf{x}_T^{\text{img}} \) and progressively generating less noisy samples \( \mathbf{x}_{T-1}^{\text{img}}, \mathbf{x}_{T-2}^{\text{img}}, \dots, \mathbf{x}_0^{\text{img}} \). The goal is to produce high-quality images by learning a denoising function \( f_\theta \) parameterized by \( \theta \) that can reverse this process.

The diffusion model training objective aims to minimize the difference between the predicted and true noise. Specifically, for each time step \( t \), the objective is to solve the following denoising problem on the image data \( \mathbf{x}^{\text{img}} \):

\[
\mathcal{L}_{\text{DM}} = \mathbb{E}_{\mathbf{x}, \epsilon \sim \mathcal{N}(\mathbf{0}, \mathbf{I}), t} \left[ \| \epsilon - f_\theta(\mathbf{x}_t^{\text{img}}, t, \mathbf{c}) \|_2^2 \right],
\]

where \( \mathbf{x}_t^{\text{img}} \) is the noisy image at time step \( t \), uniformly sampled from \( \{1, \dots, T\} \), and \( f_\theta(\mathbf{x}_t^{\text{img}}, t, \mathbf{c}) \) is the denoising function that predicts the noise added to \( \mathbf{x}_t^{\text{img}} \) conditioned on the time step \( t \) and context \( \mathbf{c} \) (often is text prompt). 

\section{X-Fusion}

In this section, we describe X-Fusion, a unified framework that adapts pretrained LLMs for vision tasks while preserving their inherent language capabilities. As illustrated in Fig.~\ref{fig:x-fusion}, X-Fusion processes both image and text inputs within a single model. To maintain the pretrained LLM’s language knowledge, we freeze its weights and introduce new trainable parameters to handle vision inputs.

\smallskip 
\noindent \textbf{Tokenizer.} Text inputs are tokenized using the original LLM's tokenizer to produce text tokens. Images are processed through a pretrained visual encoder to obtain latent representations. The encoders can be either a low-level feature compression model, like latent VAE~\cite{LDM}, or CLIP~\cite{clip}/DINO~\cite{dinov2}-like semantic encoders. These latent representations can optionally be further encoded into vision tokens using an additional trainable vision layer. In most of our experiments, we use VAE from SD1.5~\cite{LDM} as the pretrained visual encoder. After that, we use a trainable linear patchify layer with a patch size of \( 2 \times 2 \) to further compress image features into image tokens. The combined interleaved image-text tokens are then passed into X-Fusion, where the trainable parameters facilitate joint optimization for both image understanding and generation.

\smallskip 
\noindent \textbf{Dual Tower.} Let denote the tokenized input embeddings as:

\vspace{-2.5mm}
\[
\mathbf{E}^{\text{in}} = \{\mathbf{e}_1, \mathbf{e}_2, \dots, \mathbf{e}_M\}.
\]

\noindent where token \( \mathbf{e}_i \) can be either text or vision token.

To effectively process a mixture of textual and visual modalities while retaining the original language information, each layer in X-Fusion should incorporate new trainable weights alongside a frozen language layer.
Specifically, we introduce two components for each layer: A frozen text transformer block \( \mathcal{F}^{\text{txt}} \), and a trainable vision transformer block \( \mathcal{F}^{\text{img}} \).  

Both components operate on the embeddings \( \mathbf{E}^{\text{in}} \), and their respective outputs are given by:  
\[
\mathbf{H}^{\text{txt}} = \mathcal{F}^{\text{txt}}(\mathbf{E}^{\text{in}}), \quad 
\mathbf{H}^{\text{img}} = \mathcal{F}^{\text{img}}(\mathbf{E}^{\text{in}}).
\]

Intuitively, the vision layer needs to process visual tokens conditioned on texture features for generation tasks. Additionally, it also need to extract features suitable for visual understanding tasks, ensuring that the visual features can be effectively interpreted by the frozen language layer.  The outputs of the text and vision blocks are selectively combined to form the output sequence of the block:

\[
\mathbf{H}^{\text{out}} = \{\mathbf{h}_1, \mathbf{h}_2, \dots, \mathbf{h}_M\},
\]

where
\[
\mathbf{h}_i =
\begin{cases}
\mathbf{h}_i^{\text{txt}}, & \text{if } x_i \in \mathbf{x}^{\text{txt}}, \\
\mathbf{h}_i^{\text{img}}, & \text{if } x_i \in \mathbf{x}^{\text{img}}.
\end{cases}
\]

Here, \(\mathbf{h}_i^{\text{txt}} \in \mathbf{H}^{\text{txt}} \) and \( \mathbf{h}_i^{\text{img}} \in \mathbf{H}^{\text{img}} \) are the outputs of the text and vision blocks, respectively, for the \( i \)-th token. In our design, we initialize each vision block by copying the parameters in the corresponding language transformed layer, which consists of one self-attention layer, one MLP layer, and two normalization layers.

Finally, each text embedding is decoded into discrete tokens through a linear classification head. The output image feature modeling can be flexible, and could involve methods such as diffusion (L2-loss), continuous autoregressive modeling (cosine regression), or discretized autoregressive modeling (cross-entropy), depending on the choice of pretrained visual encoder and design decisions. In this paper, we primarily focus on diffusion modeling. It is important to note that the best modeling approach for new modalities is case-specific and may be explored as future work.

\noindent \textbf{X-Fuse (Optional)}
In the explanation above, we select vision tokens from the vision layers and text tokens from the text layers. Thus, in practice, we do not need to calculate the vision query tokens in the text layer, nor the text query tokens in the vision layers. This is a main design choice, as it results in the same attention FLOPs as other baseline variants shown in Fig.~\ref{fig:pipeline comparaison}, which we will discuss in the experimental section.

Optionally, we introduce an operation called X-Fuse, which sacrifices FLOPs to improve performance. Taking the text feature as an example, we also calculate the text query features in the vision layer. To fuse the text feature from both towers, we compute:

\[
\alpha * \mathbf{h}_i^{\text{txt-txt}} + \beta * \mathbf{h}_i^{\text{txt-img}}
\]
Here, the superscript indicates the source tower of the text features: "txt-txt" refers to text features from the text tower, and "txt-img" refers to text features from the vision tower. The scalars $\alpha$ and $\beta$ are learnable parameters. A similar operation can be used to fuse image features as well.

\noindent \textbf{Training.}
The final training objective combines the autoregressive loss ($\mathcal{L}_{\text{AR}}$) and the image denoising loss ($\mathcal{L}_{\text{DM}}$), with their respective weighting coefficients $\lambda_{\text{AR}}$ and $\lambda_{\text{DM}}$:

\[
\mathcal{L} = \lambda_{\text{AR}} \cdot \mathcal{L}_{\text{AR}} + \lambda_{\text{DM}} \cdot  \mathcal{L}_{\text{DM}}
\]  
We choose LLaMA-3 family~\cite{llama3} as our as pretrained LLMs for X-Fusion and use the flow matching scheduler following Stable Diffusion 3~\cite{stablediffusion_3}. We 
set $\lambda_{\text{AR}} = 0.2$ and $\lambda_{\text{DM}} = 1$, then train with AdamW optimizer ($\beta_{1}=0.9$, $\beta_{2}=0.95$), a linear warm-up scheduler, a maximum learning rate of $\textit{lr}=1\times e^{-4}$, and DeepSpeed Stage 2~\cite{deepseed} distributed training on H100 GPUs.
We train with batches of 0.8M tokens for 100k steps, processing a total of 0.08T tokens for most experiments. We extend the training to 200K steps on our X-Fusion model initialized from LLaMA-3.1-8B and report the evaluation results in Supplementary Section A. 
\noindent \textbf{Task.}
We evaluate X-Fusion's performance on: (i) image understanding and (ii) image generation, as these tasks reflect the connection between visual and textual modalities.
\begin{itemize}
    \item \textit{Image Generation}: To assess image generation, we evaluate text-to-image performance. Specifically, we generate 30K images using randomly sampled prompts from the MS-COCO~\cite{lin2014mscoco} and report image quality using FID~\cite{heusel2017fid}.
    \item \textit{Image Understanding}: To assess image understanding, we evaluate image captioning (or image-to-text) performance. Specifically, we generate captions for 30K images from the MS-COCO dataset~\cite{lin2014mscoco} and report caption quality using BLIP2-ITM~\cite{li2023blip2}.
    We also considered additional metrics, such as CIDEr~\cite{cider} and BertScore~\cite{bertscore}, however, these were either inappropriate for long captions or failed to capture meaningful changes during training. Further analysis can be found in the Supplementary.
\end{itemize}

\noindent \textbf{Data.} Otherwise stated, we sample 0.08T tokens/patches from an in-house licensed dataset. We create image-caption pairs by center-cropping and resizing images to $256 \times 256$, and pairing them with detailed captions generated by the InternVL-2.0 26B model~\cite{chen2024internvl}. These pairs are then formatted for both image-to-text tasks (serving as understanding data), and text-to-image tasks (serving as generation data).

\begin{figure*}[!h]
\centering 
\vspace{-3mm}
\includegraphics[width=0.95\textwidth]{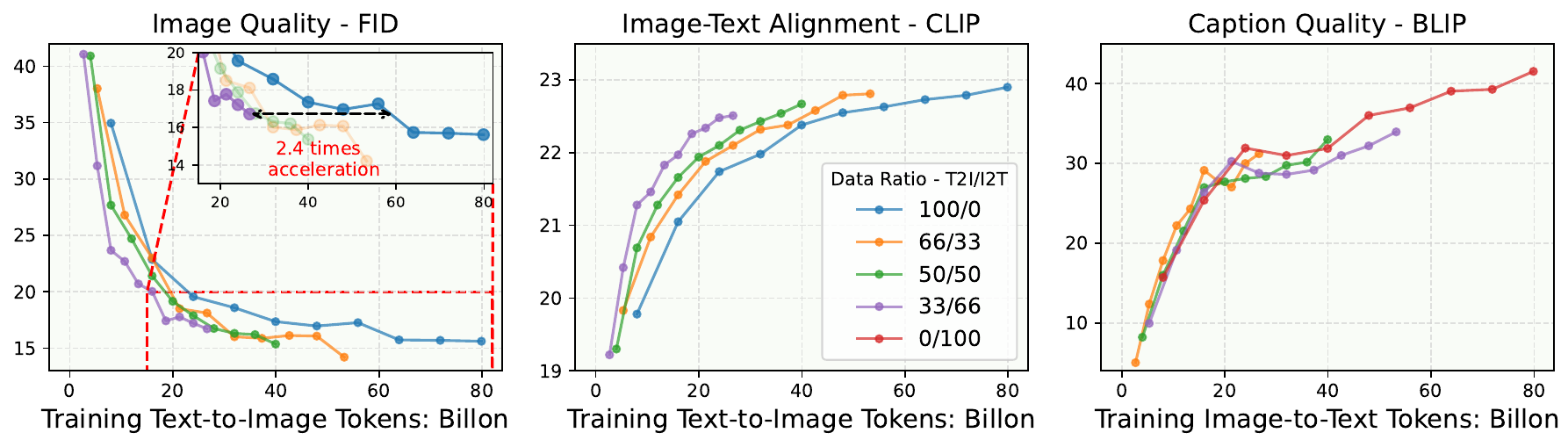}
\vspace{-3mm}
    \caption{\textbf{Performance of image generation and understanding at various data ratios.}
    Increasing visual understanding data improves visual generation performance. }
\vspace{-3mm}
\label{fig: data-ratio}
\end{figure*}
\begin{figure*}[!th]
\centering 
\includegraphics[width=0.95\textwidth]{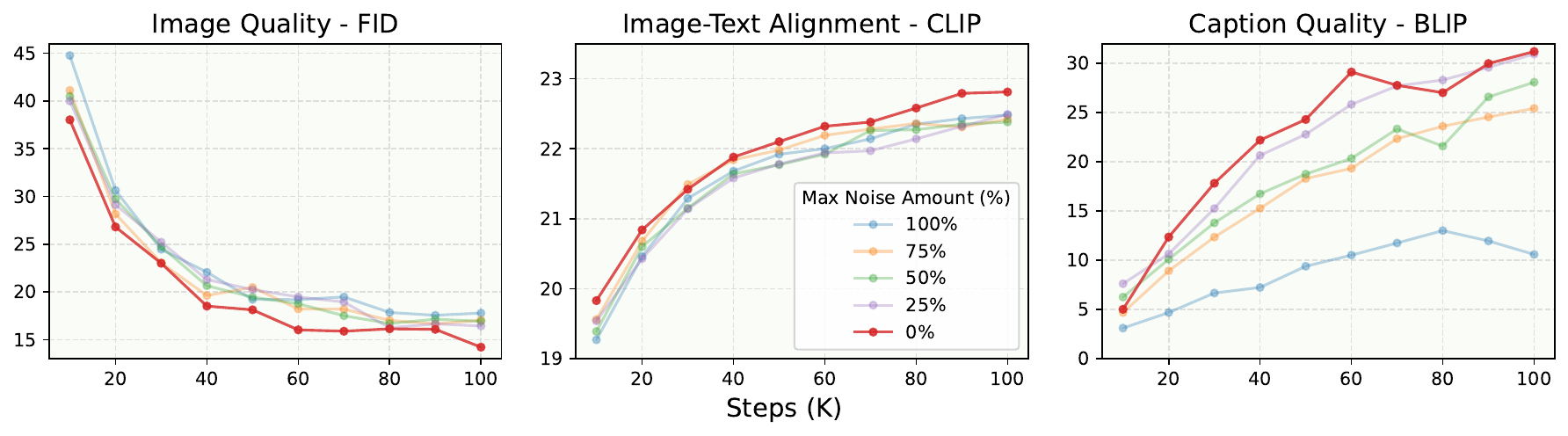}
\vspace{-4mm}
    \caption{\textbf{Performance of image generation and understanding at various noise limits in the image-to-text samples.} Providing clear images for image-to-text samples enhances visual generation and understanding simultaneously.}
\label{fig: noise-ratio}
\end{figure*}
\section{Architecture Design Choices for Adding Vision Abilities}
\label{sec:ablation_layer}

\begin{table}[t]
\small
  \centering
  \resizebox{0.45\textwidth}{!}{ 
  \begin{tabular}{@{\hspace{5pt}} l @{\hspace{5pt}} c @{\hspace{5pt}} c @{\hspace{5pt}} c @{\hspace{5pt}} c @{\hspace{5pt}}}
    \toprule
     & text only & \multicolumn{2}{c}{text2img} & img2text \\
    \cmidrule(lr){2-2} \cmidrule(lr){3-4} \cmidrule(lr){5-5}
    \textbf{Model} & \textbf{MMLU} ($\uparrow$)  & \textbf{FID ($\downarrow$)} & \textbf{CLIP ($\uparrow$)} & \textbf{BLIP}($\uparrow$) \\
    
    \midrule
    \parbox{2.5cm}{\textit{LLaMA3.2-1B}~\cite{llama3}} & 
    32.2 & --- & --- & --- \\
        \midrule
    \parbox{2.5cm}{\textit{Single Tower}} & 
    25.0 & \underline{19.10} & \underline{22.63} & 30.2 \\
    
    \parbox{2.5cm}{\textit{Gated Tower}} & 
    \textbf{32.2} & 24.51 & 21.91 & 14.5 \\
    \parbox{2.5cm}{\textit{Dual Projection}} & 
    \textbf{32.2} & 20.22 & 22.46 & \underline{30.9} \\
    \parbox{2.5cm}{\textit{Dual Tower (\textbf{Ours})}} & 
    \textbf{32.2} & \textbf{14.20} & \textbf{22.81} & \textbf{31.3} \\
    
    \bottomrule
  \end{tabular}
  }
  \caption{
  \textbf{Architecture design comparison.}
  Our dual-tower approach surpasses other baselines in image generation tasks and delivers competitive performance in image understanding, while maintaining the original language capability.
  }
  \vspace{-3mm}
  \label{tab:model_comparison}
\end{table} 

To evaluate the effectiveness of our Dual Tower design, we compare it against three alternative transformer block variants (Fig.~\ref{fig:pipeline comparaison}) that are designed for multimodal integration. Apart from the transformer blocks, all other components—including tokenizers, encoder, and decoder modules—are kept identical across configurations.

\begin{itemize}
    \item \textit{Single Tower:}
    This simple baseline uses the original language model transformer block to process both inputs directly.
    Note that this is equivalent to Transfusion~\cite{transfusion} but with pre-trained LLM as initialization (Fig.~\ref{fig:pipeline comparaison}a).
    
    \item \textit{Gated Tower:} Inspired by~\cite{alayrac2022flamingo}, we duplicate the language transformer block into a trainable ``gated block'' (Fig.~\ref{fig:pipeline comparaison}b), with both blocks taking the same input sequence:
    \[
    \mathbf{H}^{\text{txt}} = \mathcal{L}^{\text{txt}}(\mathbf{E}^{\text{in}}), \quad 
    \mathbf{H}^{\text{gate}} = \mathcal{L}^{\text{gate}}(\mathbf{E}^{\text{in}}).
    \]
    After that, the gated block output will be added to the language transformer block output, multiplied by a learnable value $\gamma$ which is initialized as 0:
    \[
    \mathbf{H}^{\text{out}} = \mathbf{H}^{\text{txt}} + \gamma *  \mathbf{H}^{\text{gate}}.
    \]
    \item \textit{Dual Projection:} We copy the language weights consisting of one self-attention and one MLP (Fig.~\ref{fig:pipeline comparaison}c). However, the data flow is different from dual-tower. First of all, based on the modality of \( \mathbf{e}_i \), we apply \textit{separate projection matrices} for query (\( \mathbf{Q} \)), key (\( \mathbf{K} \)), and value (\( \mathbf{V} \)), and then a joint attention is operated on all tokens:
    \[
    \mathbf{H}^{\text{attn}} = \text{attn}( \text{QKV}_{\text{modality}}(\mathbf{e}_i)),
    \]
    After that, the output feature from self-attention is passed through a modality-specific MLP:
    \[
    \mathbf{H}^{\text{out}} = \text{MLP}_{\text{modality}}( \mathbf{H}^{\text{attn}} ).
    \]
    This variant is similar to concurrent work~\cite{shi2024llamafusion}. However, dual-tower offers more flexibility, as the vision and language layers can be designed differently.
\end{itemize}

\noindent \textbf{FLOPs.}  Let \( N \) denote the number of text tokens and \( M \) denote the number of image tokens. For all three alternative designs, the attention complexity is \( O((N+M)^2) \). In our dual-tower design, for a fair comparison, we choose not to use the X-Fuse operation. As a result, the complexity in the vision tower is \( O(M \cdot (N+M)) \) (since we do not need query tokens for text), and the complexity in the text tower is \( O(N \cdot (N+M)) \) (since we do not need query tokens for image). In total, this results in the same complexity of \( O((N+M)^2) \). The effectiveness of X-Fuse will be presented in a later section. 

\vspace{1mm}
\noindent \textbf{Quantitative Results.} 
Tab.~\ref{tab:model_comparison} presents a comparison of different architectural designs, all are using pretrained LLaMA-3.2-1B~\cite{llama3}. Among them, the Dual Tower architecture achieves the best performance in both image generation and understanding tasks. In contrast, the Gated Layer architecture performs the weakest in both tasks, likely due to the limitations of simple addition operations. Among baselines, the Single-Tower model delivers decent performance across both tasks; however, our Dual Tower model achieves a 23\% lower FID while maintaining the same number of training parameters. More importantly, the Single-Tower model compromises the inherent knowledge of the original language model due to training on T2I and I2T tasks. To assess the model’s general knowledge, we evaluate it on the MMLU benchmark~\cite{mmmlu}, a multiple-choice test with four answer options, using a 5-shot setting.
Results show that the Single Tower model's performance drops to 25.0, which is equivalent to chance-level performance.

Dual Tower and Dual Projection share a common insight: modality-specific operations. While their text generation capabilities are nearly same, Dual Tower outperforms in image generation. This superiority can be attributed to its flexibility: The vision layers in Dual Tower can generate new QKV representations for input text features, followed by attention operations between the image and the new text features. In contrast, Dual Projection is limited to using the original language model's text QKV matrices for attention computation.
Also, it is worth noting that conceptually, Dual Tower offers greater flexibility: while our current implementation replicates the language layer as the vision layer, the vision layer does not have to be identical to the language layer, as long as it produces outputs with the same feature dimensions.
\textit{For the rest of the paper, all experiments will use Dual Tower design}.

\begin{findingbox}
\faEdit \textbf{ } Text processing flexibility affects image generation but not image understanding performance.
\end{findingbox}

\section{Effect of Data Ratios and Noise on Generation and Understanding Tasks }
We now turn to examine the effects of training data on X-Fusion's performance. In the following section, we address two key questions: (i) How does noise level affect performance in a joint training setting? (ii) Does multitask training provide mutual benefits between tasks?
\begin{figure*}[!h]
\centering 
\vspace{-3mm}
\includegraphics[width=0.9\textwidth]{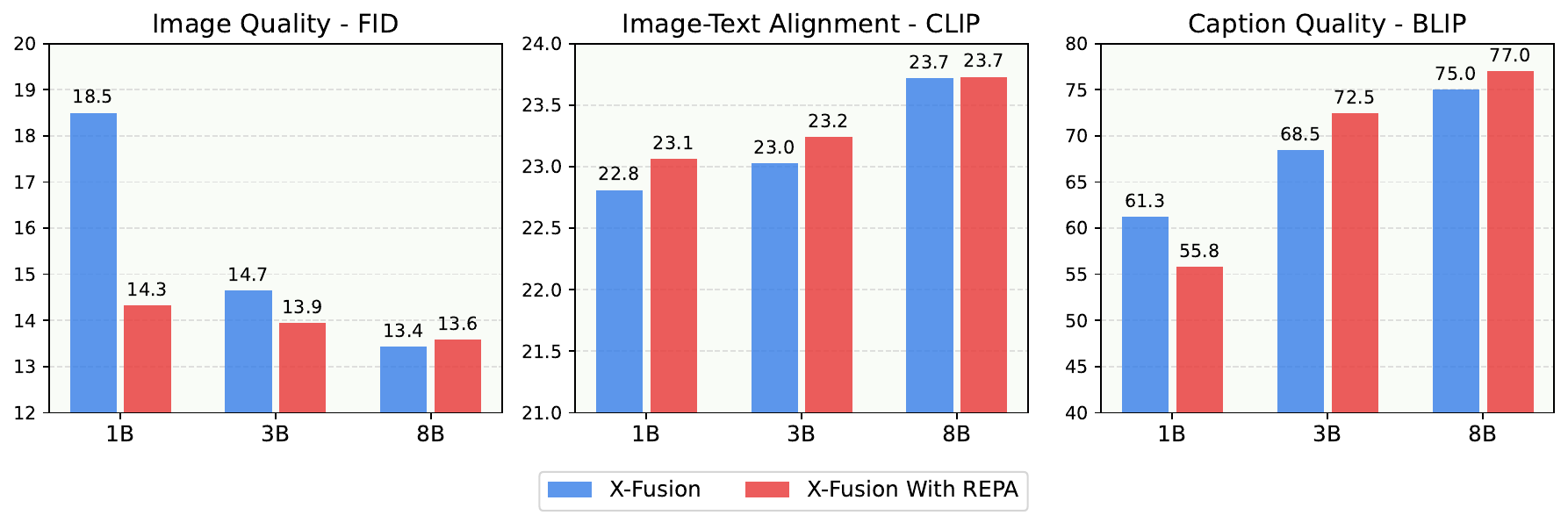}
\vspace{-3mm}
\caption{\textbf{Performance comparison of models of different sizes (1B, 3B, and 8B) with and without additional feature alignment loss.} The effectiveness of alignment diminishes as model size increases.
}
\label{fig:repa}
\vspace{-4mm}
\end{figure*}

\subsection{Effect of Noise Amount}

As a unified model, it should be able to perform both image denoising and text autoregressive tasks simultaneously on input data. However, the level of noise applied to images in image-to-text (I2T) samples remains a question. While diffusion-based image modeling benefits from noisy input for generation tasks, excessive noise can degrade visual quality and hinder image understanding. This issue was also noted by~\cite{transfusion}, where they proposed limiting the diffusion noise on I2T samples to a maximum of $t= 50\% T$ to reduce distortion for visual understanding while still can treat those noisy images as generation training data.

We argue that while this approach helps, it may not be optimal. We hypothesize that training with clean images (i.e., without adding noise to I2T samples) can lead to a stronger vision tower for image understanding, ultimately improving generation quality, although this reduces the amount of denoising data available for generation. 

To validate our hypothesis,we conducted comprehensive experiments where we systematically varied the max noise level for image understanding (I2T) tasks: 0\% (clean image), 25\%, 50\%, 75\%, and 100\% (normal generation setting).
Throughout these experiments, we maintained a 1:2 data ratio between understanding and generation tasks.
Results are provided in Fig.~\ref{fig: noise-ratio}.
As can be seen, generally the more noisy the images are, the more image understanding performance is degraded. Our proposed strategy (providing clean image for understanding) consistently achieves the best results for image understanding (2nd and 3rd col.).
Interestingly, using clean image for understanding also helps to boost performance of image generation! (1st col.). 

\begin{figure}[t]
    \centering
    \includegraphics[width=1\linewidth]{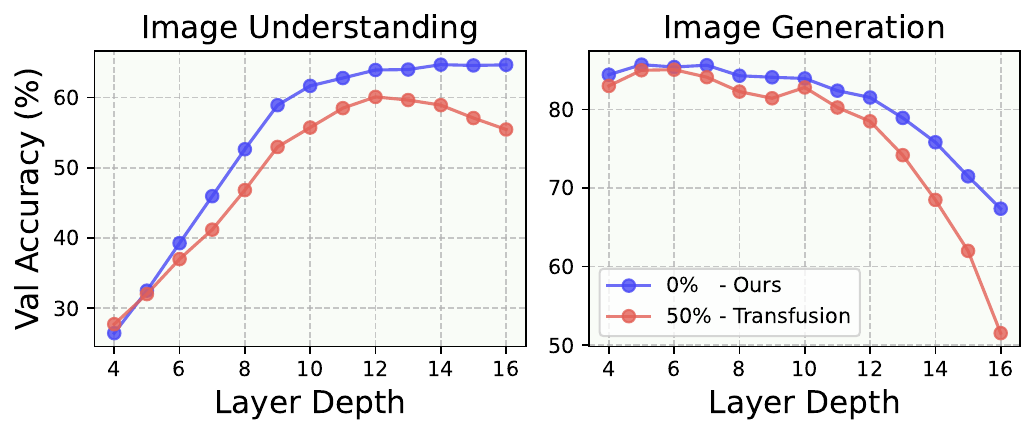}
    \vspace{-6mm}
    \caption{ \textbf{Linear Probe Results.} We use the trained model as a feature extractor and train an additional linear layer for image classification on ImageNet~\cite{imagenet}. Models trained with our training strategy constantly obtain higher feature quality.
    }

    \label{fig:noise_linear}
    \vspace{-5mm}
\end{figure}

Following \cite{repa}, we analyze the model's behavior by conducting layer-wise feature representation experiments through linear probing. We report the Top-1 accuracy on the ImageNet dataset \cite{imagenet}. As shown in Fig.~\ref{fig:noise_linear}, using clean images results in better feature representation for understanding tasks compared to noisy images (Transfusion setting). Similarly, the generative features are also superior, leading to improved generation results. We provide details of linear probing experiments in Supplementary Section A.

\begin{findingbox}
    \faEdit \textbf{ } Using clean images for visual understanding improve performance for both tasks.
\end{findingbox}

\subsection{Effect of Data Ratio}

Along with the noise addition strategy, task data ratio is also a major factor for training unified multimodal models. To investigate the synergy between visual understanding and generation from data's perspective, we kept the architecture unchanged and trained for 100k steps with a batch size of 0.8M tokens on a total of 0.08T tokens, varying the composition of text-to-image (T2I) and image-to-text (I2T) tasks.
Specifically, we begin with the training data composed entirely of T2I tasks (100\% T2I, 0\% I2T, or denoted as 100/0 for short), then progressively decrease the proportion of T2I data while increasing I2T data (i.e., 66/33, 50/50, 33/66) until the dataset consists solely of I2T tasks (0/100). Results are reported at every 10k iterations.

Fig.~\ref{fig: data-ratio} illustrates the results.
To investigate how training data from one task influences performance on other tasks, we plot performance metrics against the number of tokens observed for each specific task (e.g., first column, x-axis indicates how many image generation tokens are trained).
In another words, we want to ask: When two models are trained on an identical number of image generation tokens, but differ in their exposure to image understanding tokens, how do their performance diverge?, and vice versa.

We observe that incorporating image understanding data (I2T) improves generation quality (T2I). As the proportion of I2T data increases while keeping total T2I data fixed, generation performance consistently improves (1st and 2nd panels). In contrast, visual generation data (T2I) does not positively impact the understanding task (I2T) (third panel). Overall, there's an asymmetric relationship: understanding data benefits generation, but generation data does not enhance understanding. Based on our findings, we recommend a 66/33 (or 2:1) training ratio, which strikes a strong balance and optimizes performance across both tasks.

\begin{findingbox}
    \faEdit \textbf{ }
    Incorporating image understanding data enhances image generation performance, while adding generation data does not impact image understanding tasks.
\end{findingbox}
\begin{figure*}[ht]
\centering 
\vspace{-2mm}
\includegraphics[width=1\textwidth,page=1]{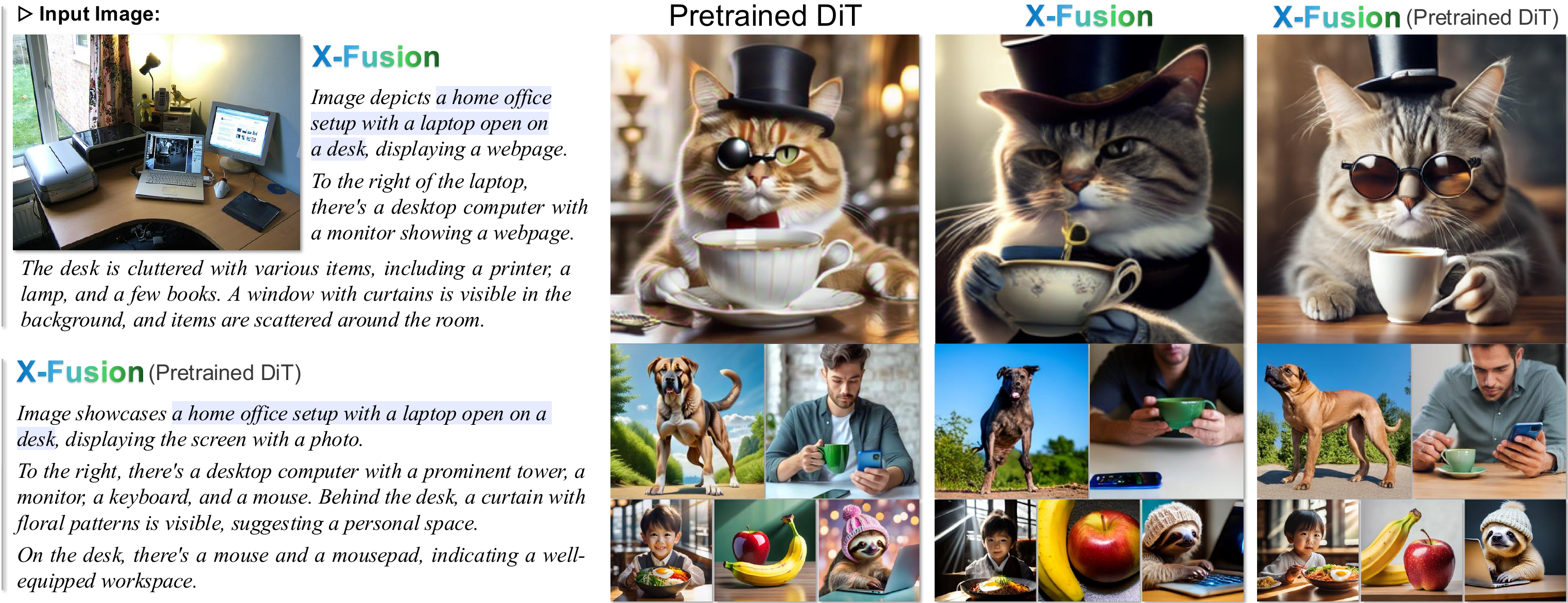}
\caption{\textbf{Qualitative comparison} between pretrained DiT model, X-Fusion(DiT) and vanilla X-Fusion on image generation and understanding task. By initializing the vision tower from pretrained text-to-image diffusion model, X-Fusion(DiT) achieves stronger image generation capability and competitive performance on image understanding compared to vanilla X-Fusion.
}
\vspace{-0.75em}

\label{fig: qual of xfusion-dit}
\end{figure*}

\section{Pretrained Representations for Vision Feature Regularization}
Inspired by~\cite{repa}, we explore whether aligning vision features in our \shortname\ with a pretrained encoder (e.g., CLIP~\cite{clip}) can improve X-Fusion’s generation and understanding.
Following~\cite{repa}, we align the vision features from the 8th layer of \shortname with the penultimate features extracted from a pretrained CLIP model.
Specifically, we take the vision feature \( \mathbf{H}^{\text{img}} \) from the dual tower and project it using a trainable linear projection \( \mathbf{W} \) to match the feature dimension of CLIP. The alignment loss is then computed as the cosine distance between the projected feature and the CLIP feature, encouraging our intermediate vision features to closely resemble those of CLIP.
The regularization loss is defined as
$\mathcal{L}_{\text{align}} = 1 - \text{cos}\left(\mathbf{W} \mathbf{H}^{\text{img}}, \mathbf{H}^{\text{CLIP}}\right)$, where \( \mathbf{H}^{\text{CLIP}} \) is the vision feature extracted from the pretrained CLIP encoder, and \( \text{cos}(\cdot, \cdot) \) represents the cosine similarity.

To evaluate effectiveness and scalability, we tested this approach on our dual-tower model with three different base model sizes: 1B, 3B, and 8B. We use $\lambda_{\text{AR}}=0.5$ in this ablation study. 
As Fig.~\ref{fig:repa} shows, alignment accelerates training and improves performance but with two key findings:
(1) Its impact diminishes with model size, even slightly degrading 8B model performance at 100k iterations. (2) Regardless of size, alignment loss imposes a performance ceiling, likely set by the external encoder’s representational power.
These findings highlight that aligning with external representations is particularly beneficial for smaller models, but the benefit may be less for larger models.
Future work could explore whether using a more powerful external encoder (e.g., a larger or improved vision backbone) might further push this performance boundary.

\begin{findingbox}
    \faEdit \textbf{ } Aligning with pre-trained vision features (i.e., CLIP) improves performance for smaller models but has significantly less impact on larger ones.
\end{findingbox}

\begin{figure}[ht]
\centering 
\includegraphics[width=1\linewidth]{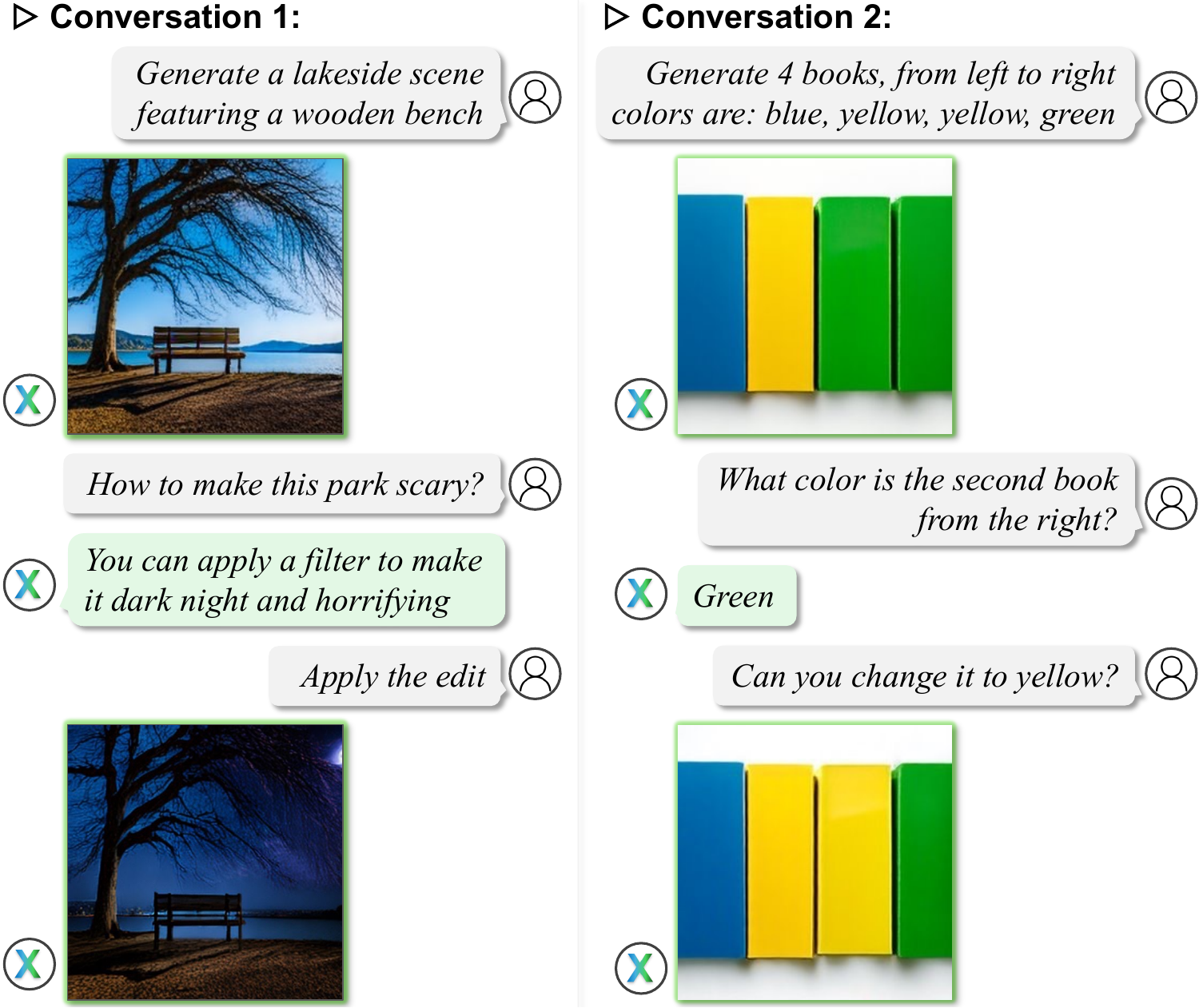}
   \vspace{-5mm}
\caption{\textbf{Interactive Generation.} Our X-Fusion model can follow user instructions to understand, generate, or edit images. }
\vspace{-1em}
\label{fig:multi-turn generation}
\end{figure}

\begin{figure*}[ht]
\centering 
\includegraphics[width=0.99\textwidth]{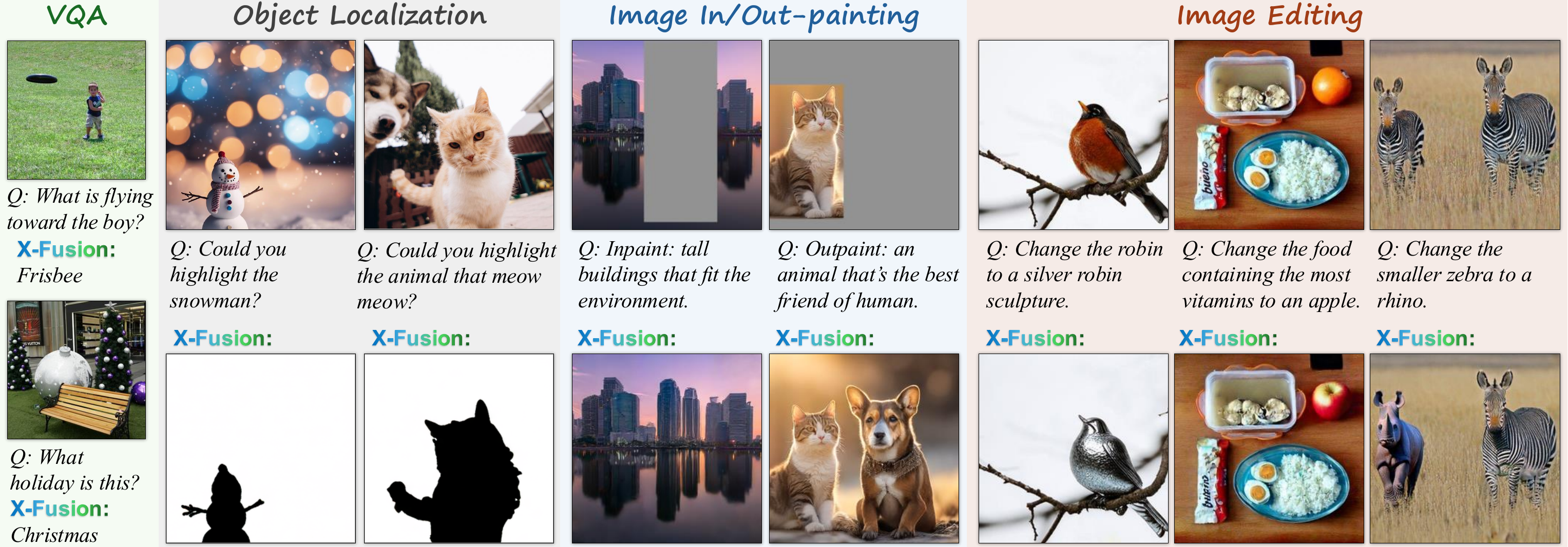}
\vspace{-1mm}
\caption{\textbf{Qualitative results} of fine-tuned X-Fusion model on downstream tasks including: visual question answering~(VQA), image editing, localization, and in/out-painting tasks. 
}
\vspace{-1em}
\label{fig: image-edit}
\end{figure*}

\section{Extension of X-Fusion}
In this section, we introduce three extensions of X-Fusion, including X-Fuse layer, transferring from pretraind diffusion models, and finetune for downstream tasks.


\begin{figure}[ht]
\centering 
\includegraphics[width=0.98\linewidth]{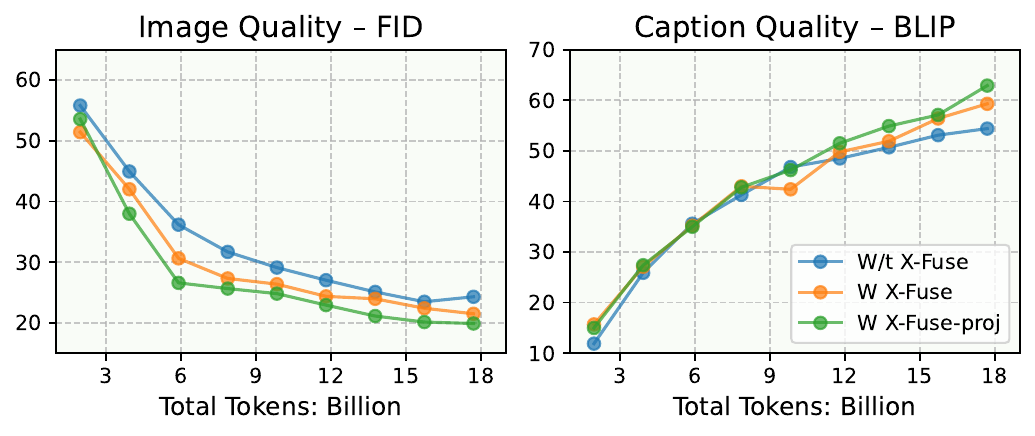}
\vspace{-0.5em}
\caption{\textbf{Ablation: X-Fuse layer}. Our X-Fusion model with the X-Fuse layer outperforms the baseline X-Fusion model on image generation and understanding tasks.  
}
\label{fig: exp xfuse}
\vspace{-1.0em}
\end{figure}

\subsection{Effectiveness of X-Fuse Layer}
So far, our study is conducted on a dual-tower architecture without the X-Fuse operation, maintaining the same FLOPs as other baseline designs. In Sec~\ref{sec:ablation_layer}, we further propose the X-Fuse operation, which merges features from both towers to trade increased FLOPs for improved performance. We conduct an ablation study to evaluate this design, and the results are shown in Fig~\ref{fig: exp xfuse}. As illustrated, applying the X-Fuse operation leads to improvements in both image generation and understanding capabilities.

\subsection{Transfer from Pretrained Diffusion Model}
While X-Fusion successfully kept its language generation capability, its image generation capability still needs to be trained from scratch. One solution is to transfer the knowledge from existing image generation models. In our dual-tower design, each block in the language and vision tower processes the entire feature sequence independently, therefore allowing non-identical block designs in both towers. With this advantage, we could transfer the image generation capability from a large-scale pretrained diffusion model that uses diffusion transformers~\cite{peebles2023dit,stablediffusion_3}.

We train a variation of X-Fusion using Llama3.1-8B as the language tower and an in-house pretrained text-to-image DiT model as its vision tower, notated as X-Fusion(Pretrained DiT), using the same training recipe in the previous section for 50K iterations. To align the feature dimension in both towers, we add linear projection layers within the X-Fuse layer. Figure~\ref{fig: exp xfuse} shows that this operation further enhances the model's capability. We also empirically find that the introduction of the X-Fuse layer could accelerate the convergence when initializing from a pretrained diffusion model. Figure~\ref{fig: qual of xfusion-dit} qualitatively compares the image generation and image understanding performance between the pre-trained DiT model, X-Fusion(Pretrained DiT), and vanilla X-Fusion-8B models. X-Fusion(Pretrained DiT) obtained stronger image generation capability and on-par image understanding performance compared to the vanilla X-Fusion.


\subsection{Fine-tune X-Fusion}
Our X-Fusion model has been pre-trained on T2I and I2T tasks and has achieved strong cross-modal performance. Can X-Fusion extend its capabilities to other downstream vision-and-language tasks? We fine-tuned our model on four tasks simultaneously—including image editing, localization, outpainting, and Visual Question Answering (\textbf{VQA})—using internal datasets for 20k training steps. In Figure~\ref{fig: image-edit}, we demonstrate that our unified X-Fusion model can handle multiple tasks without creating task-specific models or weights. We evaluated image-editing performance on publicly available datasets, including PIE-Bench~\cite{ju2023pie_bench} and SmartEdit~\cite{huang2024smartedit}. Notably, X-Fusion showcased strong instruction-based editing capability when challenged to select between multiple objects—such as “smaller zebra” and “food containing the most vitamins.” Furthermore, driven by X-Fusion’s generalized capability across vision-and-language tasks, we unlock new interactive vision–language applications by enabling a single model to both generate, understand, and edit images with natural-language instructions from users, as illustrated in Figure~\ref{fig:multi-turn generation}.

\section{Conclusion}

This paper introduces X-Fusion, a novel framework for adapting pretrained Large Language Models to new modalities (e.g., vision) while retaining their original language capabilities.
We propose a Dual Tower architecture in which language weights remain frozen, while visual features are processed via a trainable vision tower with separate weights.
Our experimental results demonstrate that this Dual Tower approach outperforms other architectural variants in both image understanding and image generation tasks.
Alongside this novel architecture, we provide a systematically comprehensive set of ablation studies that offer valuable insights from a data perspective. Our findings reveal that: (i) incorporating understanding-focused data improves generation performance, (ii) reducing noise in image data enhances overall results, and (iii) feature alignment benefits primarily smaller models.
We hope our paper will step forward building an unified Large Multimodal Models in a more effiecient way.

\newpage
{
    \small
    \bibliographystyle{unsrt}
    \bibliography{reference}
}
\

\newpage
\twocolumn[
\begin{center}
{\Large \bf X-Fusion: Introducing New Modality to Frozen Large Language Models  \\(Supplementary Material)\par}
{\large
\lineskip .5em
\begin{tabular}[t]{c}
\end{tabular}
\par}
\vskip .5em
\end{center}
]


\setcounter{figure}{0}
\setcounter{table}{0}
\setcounter{section}{0}
\renewcommand{\thefigure}{\Alph{figure}}
\renewcommand{\thesection}{\Alph{section}}
\renewcommand{\thetable}{\Alph{table}}

In the supplementary material, we present implementation details (Section~\ref{sec:implementation}) and additional experiment results and analysis(Section~\ref{sec:exp_resutls}). We also discuss the  societal impact of our method (Section~\ref{sec:impact}) and limitations (Section~\ref{sec:limitation}). For sections and figures, we use numbers (\eg, Sec.\ 1) to refer to the main paper and capital letters (\eg, Sec.\ A) to refer to this supplement. We hope this document complements the main paper.


\section{Implementation Details}
\label{sec:implementation}
In this section, we present the implementation details, including the model, experiments, and the choice of evaluation metrics.

\subsection{Model Details}
Our X-Fusion uses the pre-trained VAE model with the compression ratio of 4 from Stable Diffusion~\cite{stablediffusion} and follows the flow matching training from Stable Diffusion 3. 
During inference, we use the classifier-free guidance scale of 5.5 as we empirically find that it provides optimal visual results.

\subsection{Caption Quality Evaluation Metric}
In our main paper, we use pairs of images and long captions (generated from InternVL~\cite{chen2024internvl}) as training data for both text-to-image (T2I) and image-to-text (I2T) tasks. The use of long captions encourages the model to generate more detailed images, while also forcing the model captures fine-grained semantic details for understanding tasks. For generation evaluation, we use standard metrics like FID~\cite{heusel2017fid} and CLIP scores~\cite{clip}. However, we discovered that the CIDEr score~\cite{cider}, commonly used to evaluate caption quality in recent papers such as Chameleon and Transfusion, is not suitable for long captions. Therefore, it is not an ideal metric for monitoring our ablation results.

\begin{figure*}[!h]
\centering 
\includegraphics[width=0.95\textwidth]{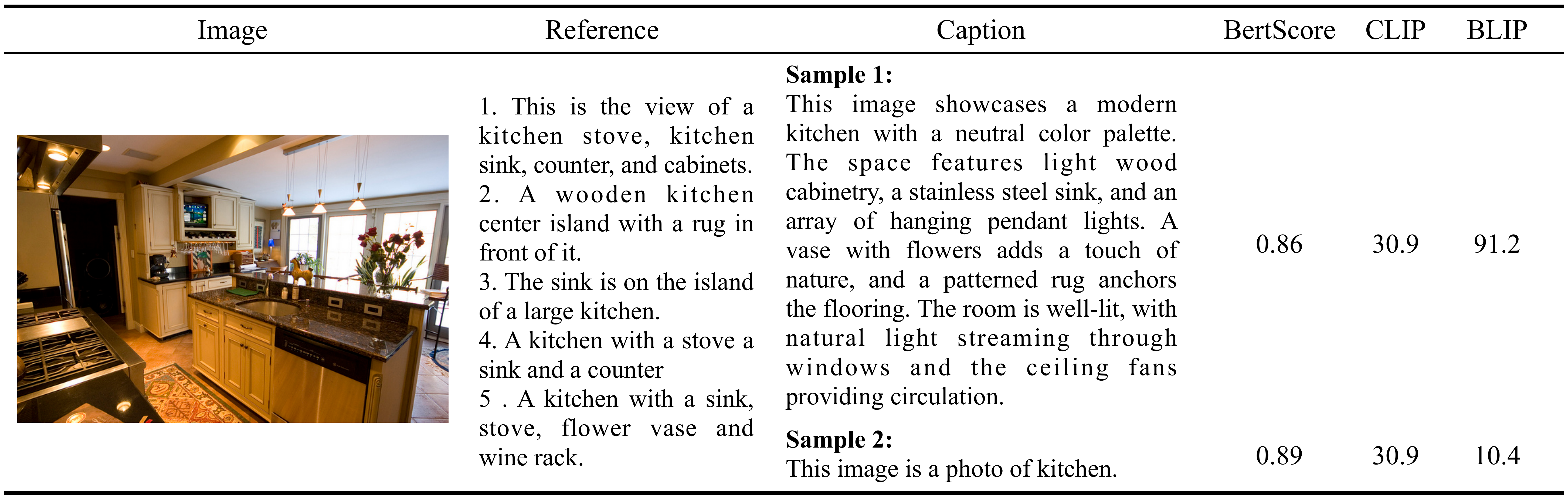}
\vspace{-1mm}
\caption{\textbf{Comparison of Different Evaluation Metrics.} The BLIP score effectively differentiates between the two captions with varying levels of detail, whereas the other metrics do not.}
\vspace{3mm}
\label{fig: cap_vis}
\end{figure*}

The CIDEr score, which relies on n-gram overlap, tends to give lower scores for long captions. This is because COCO ground truth captions are short, and even accurate long captions may have minimal n-gram overlap, leading to a lower score. Our evaluation shows that the CIDEr score between human-written COCO short captions and InternVL long captions is only 1e-5, an insignificantly low value. We also evaluate other metrics, such as BertScore~\cite{bertscore} and CLIP score~\cite{clip}. However, neither is well-suited for assessing long captions. Figure~\ref{fig: cap_vis} presents qualitative examples where the reference caption is a human-written COCO ground truth caption. Among the generated captions, Sample 1 is a long and informative caption, while Sample 2 is short and of low quality. As shown in the figure, both BertScore and CLIP score assign relatively high scores to these two vastly different captions, making it difficult to gauge the quality of training progress. In contrast, the BLIP score effectively differentiates between the two captions, making it a more suitable evaluation metric.




\subsection{Linear Probing Experiments}
We followed previous literature~\cite{repa,Chen2024DeconstructingDD} for the details of the linear probing experiment settings. We use the trained Dual Tower model as the feature extractor and train an additional classification head along with a parameter-free normalization layer. Specifically, we extract the image representation from the vision tower instead of the language tower. We used a batch size of 16,384 with the Adam optimizer without weight decay and set the learning rate to $6.4~\times10^{-3}$. We report the top-1 accuracy on the validation dataset.

As the objective of the linear probing experiments is to examine the quality of the learned visual features, we consider two settings for extracting the visual representation when the model is used for visual understanding tasks and visual generation tasks. For the understanding mode, we provide clear images without noise and set the corresponding timestep to $t=0$. For the generation mode, since X-Fusion is partially trained with text-to-image generation tasks, we add a text prompt preceding the noisy image and set the corresponding timestep to $t=20$.


\begin{table*}[]
\centering
\begin{tabular}{lcccc}
\toprule[1.5pt] 
                         &              & \textbf{Language} & \textbf{Image Understanding} & \textbf{Image Generation} \\

\textbf{Method}                   & \textbf{Base LLM}    & \textbf{MMLU}~$\uparrow$     & \textbf{COCO BLIP}~$\uparrow$        & \textbf{COCO FID}~$\downarrow$         \\
\midrule
\color[HTML]{9b9b9b}\textbf{Language Generation Only} &              &          &                  &                  \\
\color[HTML]{9b9b9b}Llama2 7B                &              & \color[HTML]{9b9b9b}45.3     & \color[HTML]{9b9b9b}-                & \color[HTML]{9b9b9b}-                \\
\color[HTML]{9b9b9b}LLaMA-3.1 8B             &              & \color[HTML]{9b9b9b}66.7     & \color[HTML]{9b9b9b}-                & \color[HTML]{9b9b9b}-                \\
\color[HTML]{9b9b9b}LLaMA-3.2-Vision 11B             &              & \color[HTML]{9b9b9b}66.7     & \color[HTML]{9b9b9b}82.3                & \color[HTML]{9b9b9b}-                \\
\color[HTML]{9b9b9b}InternVL2.0-26B             &              & \color[HTML]{9b9b9b} -     & \color[HTML]{9b9b9b}81.1              & \color[HTML]{9b9b9b} -                \\
\color[HTML]{9b9b9b}\textbf{Visual Generation Only}   &              &          &                  &                  \\
\color[HTML]{9b9b9b}Stable Diffusion         &              & \color[HTML]{9b9b9b}-        & \color[HTML]{9b9b9b}-                & \color[HTML]{9b9b9b}9.6              \\
\midrule
\textbf{Unified Models}           &              &          &                  &                  \\
Emu-3                    &              & 35.3        & 79.6             & 12.8             \\
Janus                    & DeepSeek1.3B & -        & 70.8              & 8.5              \\
Chameleon-7B             &              & 52.1     & 54.1              & 26.7             \\
Show-O                   & Phi-1.5B     & -        & -              & 9.2             \\
Transfusion              &              & -        & -                & 6.7              \\
LLaVAFusion              & LLaVA-Next8B & -        & -                & 8.2              \\
MetaMorph                & LLaMA-3.1 8B & -        & -                & 11.8             \\
X-Fusion (Ours)          & LLaMA-3.1 8B & 66.7     & 80.0             & 11.5             \\       
\bottomrule[1.5pt] 
\end{tabular}
\caption{\textbf{Quantitative Comparison with State-of-the-Art Models.} X-Fusion achieves competitive performance compared to other leading unified models. For the image understanding task, we prompt open-sourced models to generate detailed captions.}
\label{table:xfusion_model}
\end{table*}

\section{Additional Experiments Results}
\label{sec:exp_resutls}

\begin{figure*}[!h]
\centering 
    \vspace{13mm}
\includegraphics[width=1.0\textwidth]{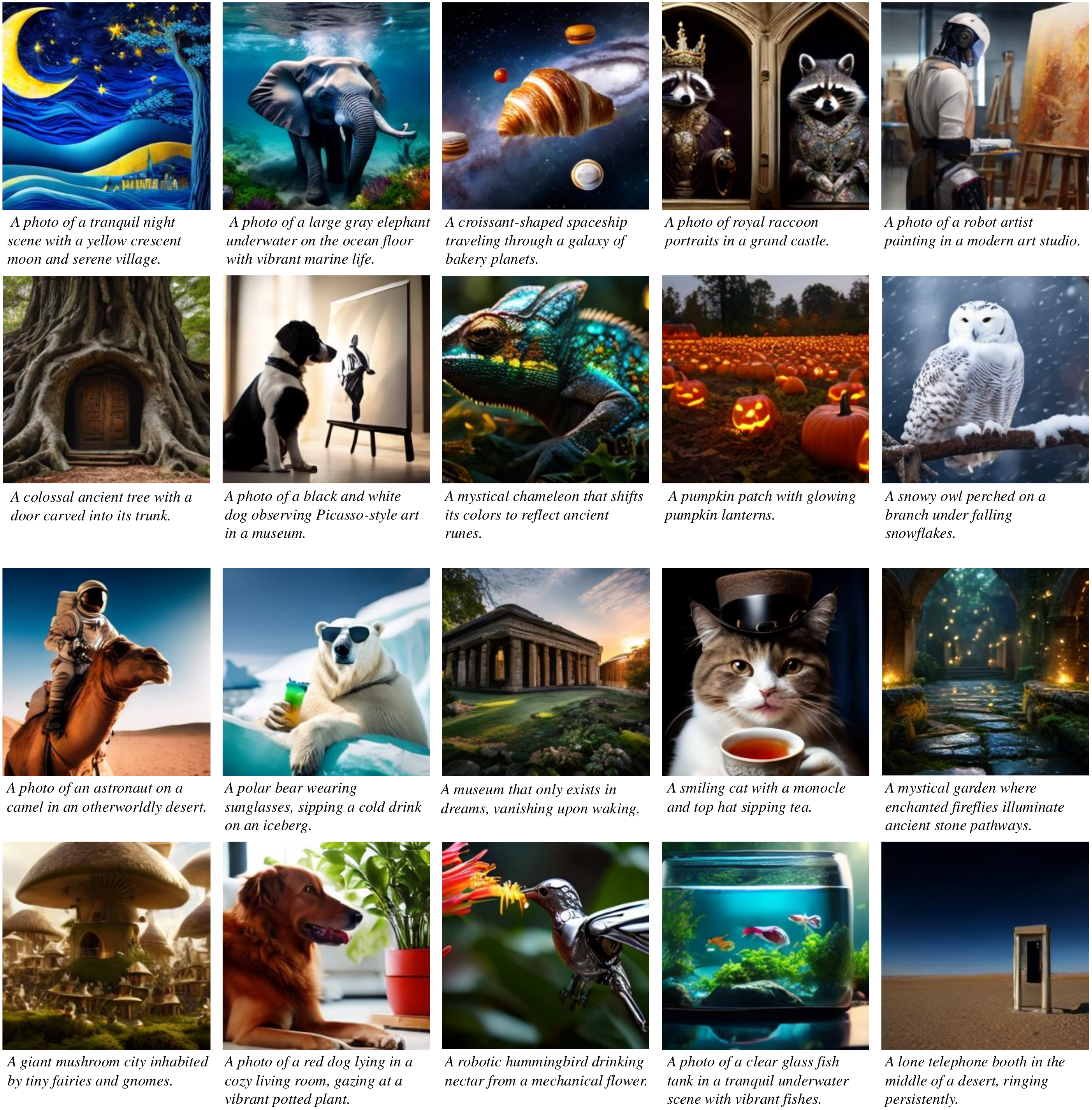}
   \vspace{-3mm}
    \caption{\textbf{Additional qualitative results of X-Fusion model.} Our generation samples demonstrated good visual quality and achieved great image-text alignment.}
    \vspace{10mm}
\label{fig: qualitative results generation}
\end{figure*}

\begin{figure*}[!h]
\centering 
\includegraphics[width=1.0\textwidth]{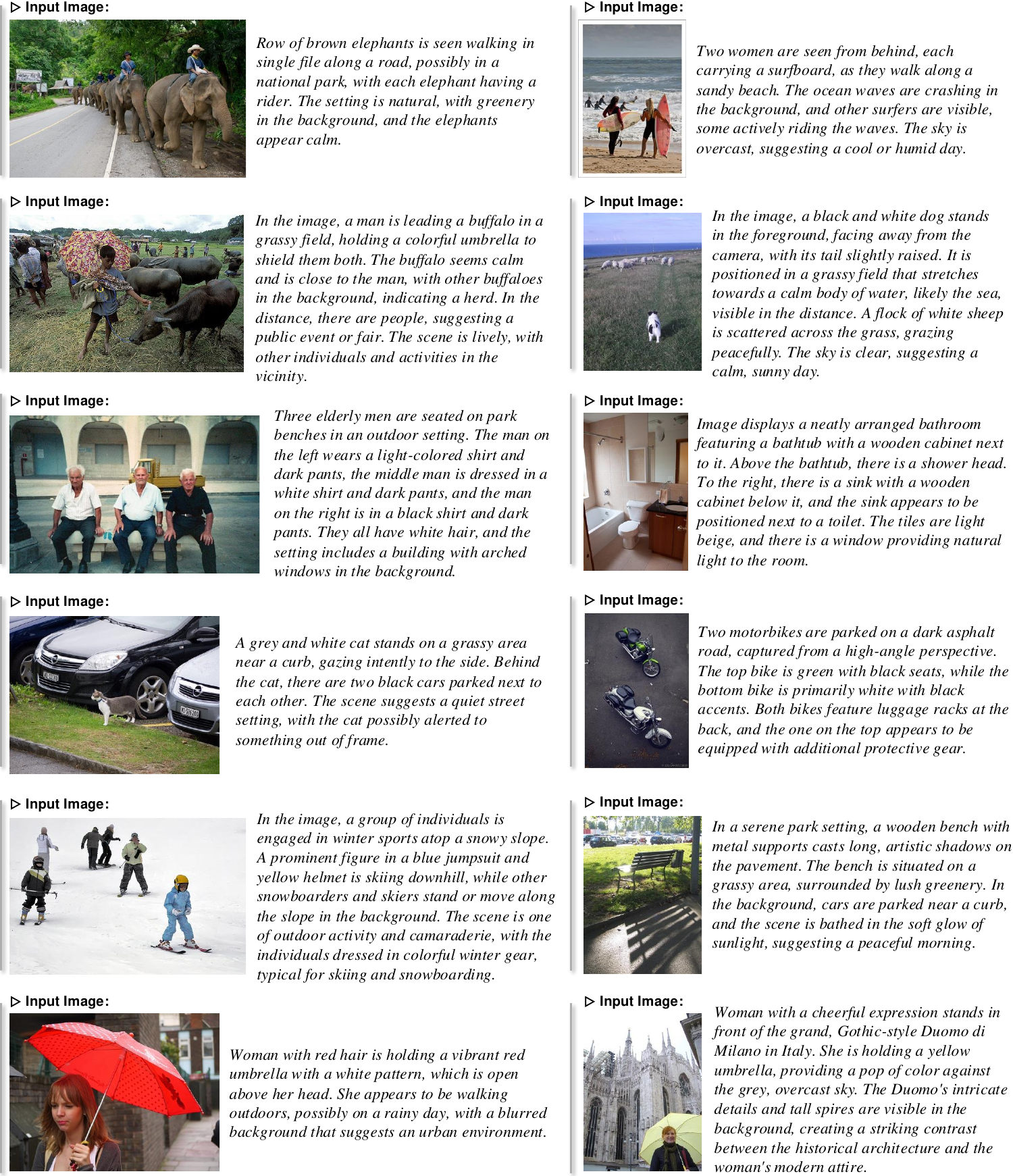}
   \vspace{-3mm}
    \caption{\textbf{Additional qualitative results of X-Fusion model.} Our generation captions demonstrated good text quality and achieved great image-text alignment.}
\label{fig: qualitative results caption}
\end{figure*}

In the main paper, we systematically explored the architectural design choices, data types, and other training factors to identify an effective approach for training a unified model with a frozen language model. Our experiments primarily focused on 1B parameter models. In this supplementary material, we extend our study by training the model with the LLaMA-3.1-8B architecture using our final training recipe. Specifically, we employ a Dual Tower design, with a 2:1 data ratio for generation and understanding tasks. Additionally, we ensure that image-to-text samples are created with cleaned images and without feature regularization from an external encoder. The model is trained with 0.8M tokens per batch for 200k iterations. Table~\ref{table:xfusion_model} compares our model's performance with other state-of-the-art unified models. Our model achieves comparable captioning and image generation quality while maintaining its original language capabilities.
We also provide additional qualitative results for image generation and image understanding in Figure~\ref{fig: qualitative results generation} and \ref{fig: qualitative results caption}, respectively. As shown in Figure~\ref{fig: qualitative results generation}, our X-Fusion achieved great visual quality and image-text alignment. Also, Figure~\ref{fig: qualitative results caption} illustrates the strong performance of X-Fusion for visual understanding. 
\begin{figure}[ht]
\centering 
\includegraphics[width=0.49\textwidth]{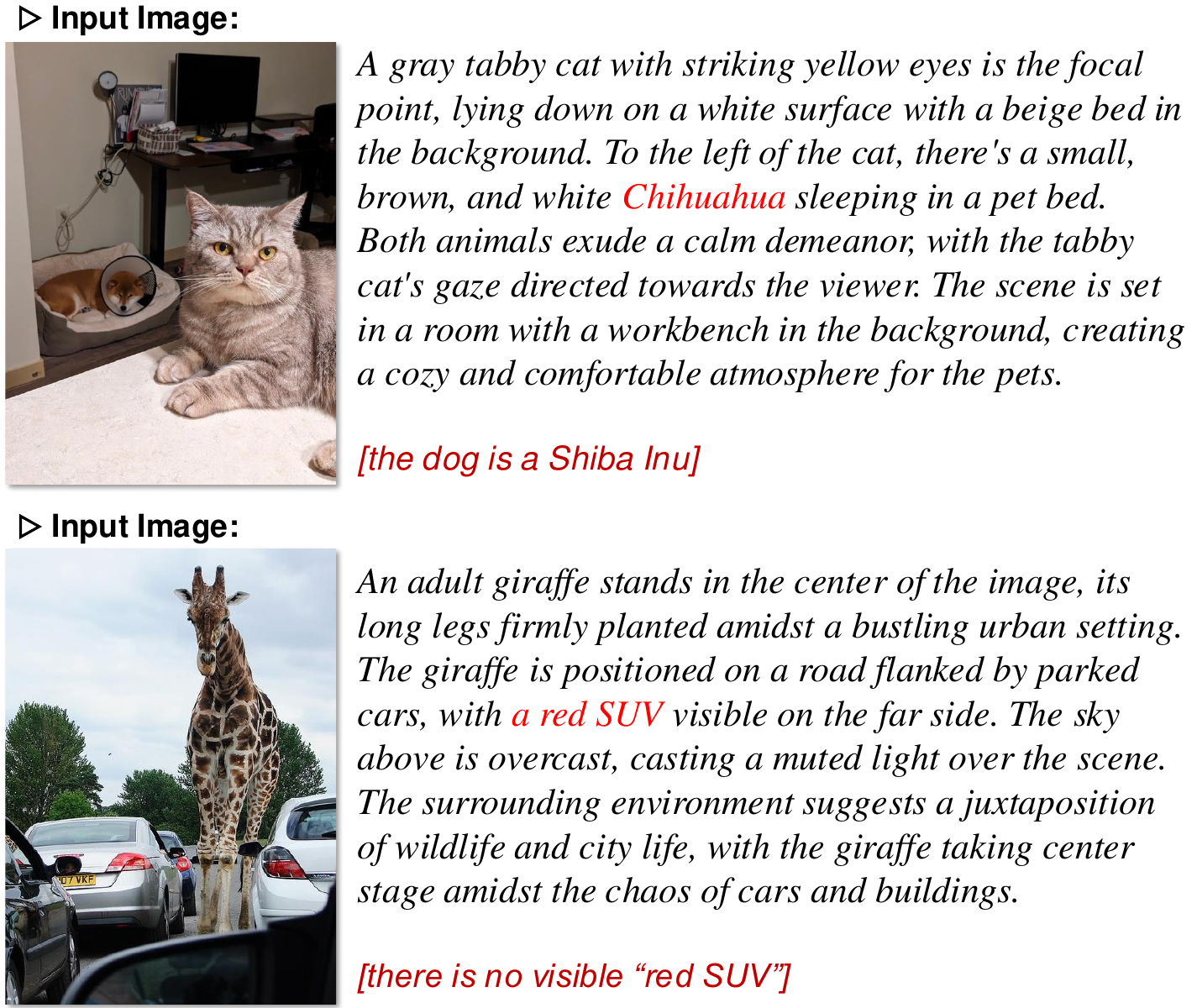}
   \vspace{-3mm}
\caption{\textbf{Limitations.} Similar to other large multimodal models, our model is prone to hallucinating in its generated concepts.}
\label{fig:limitation}
\end{figure}
\section{Social Impact}
\label{sec:impact}

Our work presents an efficient approach to integrating new modalities into frozen LLMs, enabling both image understanding and generation. By systematically studying architectural and data-centric design choices, we provide insights that enhance the scalability and efficiency of multimodal learning, reducing computational costs and making such models more accessible. This has potential applications in assistive AI, creative content generation, and vision-language understanding, benefiting areas like accessibility and education.

However, multimodal AI also raises ethical concerns, including bias in training data, misinterpretation of images, and potential misuse of generated content. To ensure responsible deployment, future work should focus on dataset fairness, robustness evaluations, and ethical safeguards. Our research contributes to the development of scalable and responsible multimodal AI, promoting more adaptable and efficient vision-language models.

\section{Limitation}
\label{sec:limitation}

X-Fusion is not without limitations. First, like other vision-language models, it is prone to hallucinations, occasionally generating inaccurate or misleading outputs (Fig.~\ref{fig:limitation}). This may be due to the model operating in the latent space of LDM~\cite{LDM}, which might not accurately capture fine details. A potential solution is to use a more robust latent VAE encoder. Second, although our model has shown promising results, there is still room for improvement in image quality, such as by training with higher-resolution images. Third, while X-Fusion's vision tower can be designed to be different from its language tower to control the number of new learnable parameters, vanilla X-Fusion doubles the number of model parameters, thereby reducing training efficiency compared to Transfusion~\cite{transfusion}. 
Nevertheless, this paper serves as a study that explores and analyzes various factors, including architecture and training strategies for unified MLLM training, offering valuable insights for future research.

\end{document}